%% file: main.tex
\documentclass[10pt,twocolumn,letterpaper]{article}

\usepackage{amsfonts}
\usepackage{dsfont}
\usepackage{cvpr} 
\usepackage[accsupp]{axessibility}

\input{preamble}
\definecolor{cvprblue}{rgb}{0.21,0.49,0.74}
\usepackage[pagebackref,breaklinks,colorlinks,
            linkcolor=cvprblue,
            citecolor=cvprblue,
            filecolor=cvprblue,
            urlcolor=.]{hyperref}
\usepackage{multirow}
\usepackage{tcolorbox}

\def\paperID{xxxxx} 
\def\confName{CVPR}
\def\confYear{2026}

\definecolor{rosepink}{RGB}{255, 102, 204}

\title{
  \raisebox{-0.245\height}{\includegraphics[width=20pt]{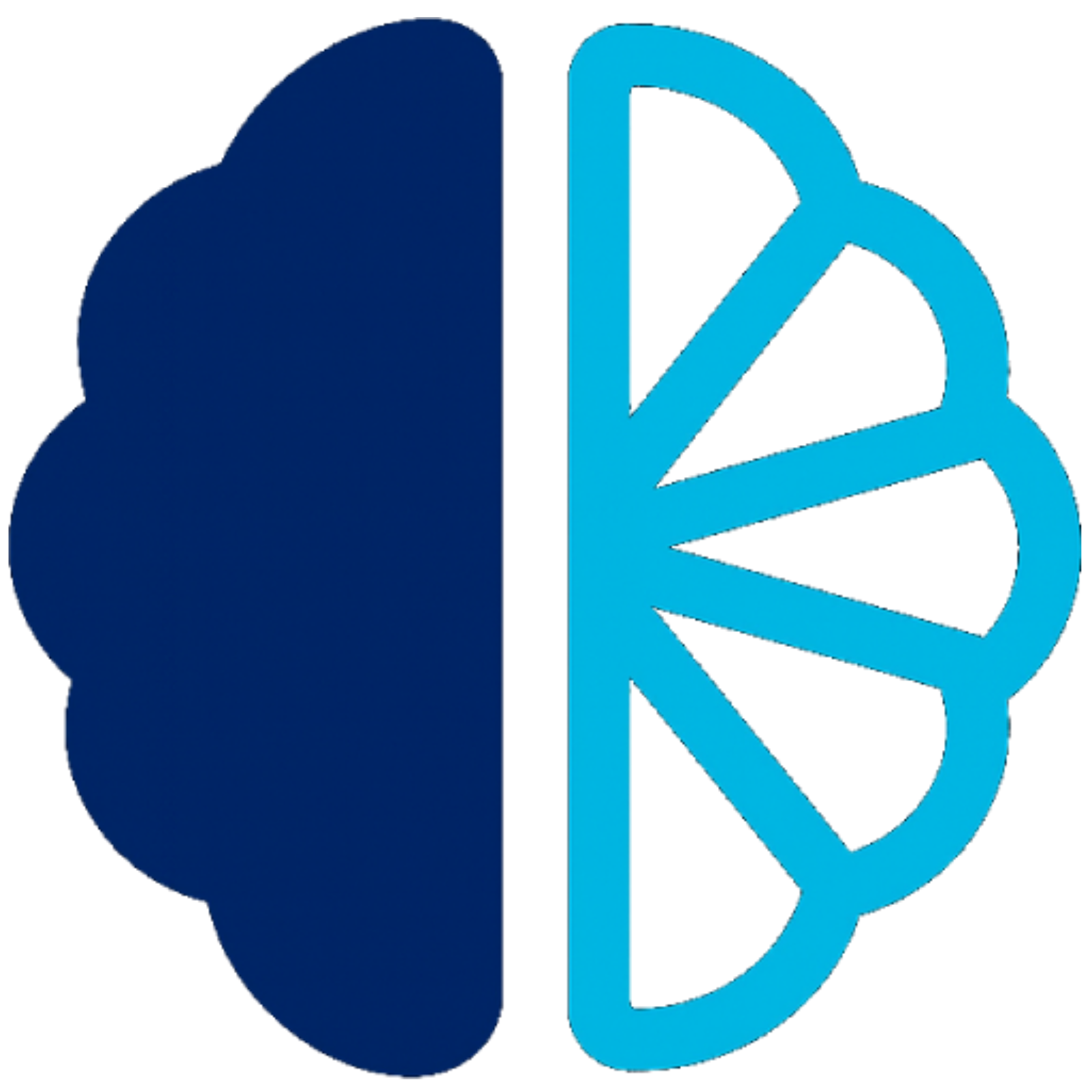}}~Progress-Think: Semantic Progress Reasoning for Vision-Language Navigation
}

\author{
Shuo Wang$^{1,3,}$\thanks{This work was done while Shuo Wang was a Research Intern with Horizon Robotics.} ,
Yucheng Wang$^{3,\dagger}$,
Guoxin Lian$^{1,3}$,
Yongcai Wang$^{1,\ddagger}$,
Maiyue Chen$^{3}$,\\
Kaihui Wang$^{3}$,
Bo Zhang$^{3}$,
Zhizhong Su$^{3}$,
Yutian Zhou$^{1}$,
Wanting Li$^{1}$,
Deying Li$^{1}$,
Zhaoxin Fan$^{2,\ddagger}$\\
[2mm]
\normalsize{
$^1$School of Information, Renmin University of China} \quad \\
\normalsize{
$^2$Beijing Advanced Innovation Center for Future Blockchain and Privacy Computing} \quad \\
\normalsize{
$^3$Horizon Robotics} \quad \\
\normalsize{
$^\dagger$Project leader} \quad
\normalsize{
$^\ddagger$Corresponding authors}\\
\small \href{https://horizonrobotics.github.io/robot_lab/progress-think}{\textcolor{rosepink}{\url{https://horizonrobotics.github.io/robot_lab/progress-think}}}
}

\begin{document}
\maketitle

\input{sec/0_abstract}

\input{sec/1_intro}
\input{sec/2_method}
\input{sec/3_experiment}

{
    \small
    \bibliographystyle{ieeenat_fullname}
    \bibliography{main}
}



\end{document}

%% file: sec/0_abstract.tex
\begin{abstract}
Vision-Language Navigation requires agents to act coherently over long horizons by understanding not only local visual context but also how far they have advanced within a multi-step instruction.
However, recent Vision-Language-Action models focus on direct action prediction and earlier progress methods predict numeric achievements; both overlook the monotonic co-progression property of the observation and instruction sequences.
Building on this insight, Progress-Think introduces semantic progress reasoning, predicting instruction-style progress from visual observations to enable more accurate navigation.
To achieve this without expensive annotations, we propose a three-stage framework.
In the initial stage, Self-Aligned Progress Pretraining bootstraps a reasoning module via a novel differentiable alignment between visual history and instruction prefixes.
Then, Progress-Guided Policy Pretraining injects learned progress states into the navigation context, guiding the policy toward consistent actions.
Finally, Progress-Policy Co-Finetuning jointly optimizes both modules with tailored progress-aware reinforcement objectives.
Experiments on R2R-CE and RxR-CE show state-of-the-art success and efficiency, demonstrating that semantic progress yields a more consistent representation of navigation advancement.
\end{abstract}

%% file: sec/1_intro.tex
\section{Introduction}
\label{sec:intro}

Vision-Language Navigation (VLN) in continuous environments \cite{wu2024vision,anderson2018vision,gu2022vision,dai2026thinkmatter} is a long-horizon multimodal reasoning task requiring agents to interpret instructions and act coherently in complex scenes. VLN demands consistent reasoning over extended visual observations, stepwise linguistic goals, and actions. This requires not only understanding local visual context but also recognizing how far the agent has advanced semantically by aligning accumulated observations with the instruction structure. However, existing VLN approaches show a clear gap in stepwise semantic progress alignment, leading agents to lack awareness of their semantic advancement within the instruction.

\begin{figure}[t]
    \centering
\includegraphics[width=1.0\linewidth]{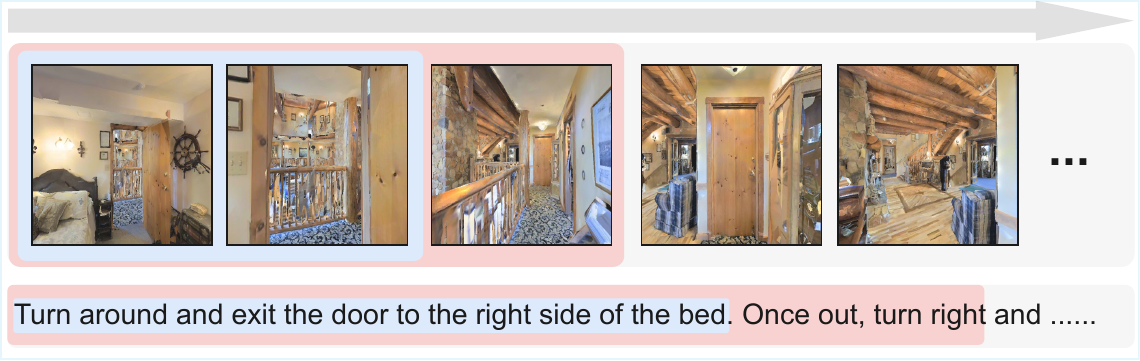}
    \caption{Our key structural insight in VLN: visual observations and instruction semantics exhibit monotonic co-progression. As observations accumulate (top), the aligned instruction prefix extends monotonically over time (bottom), with later progress (red) consistently building on earlier progress (blue). }
    \label{fig:progress}
\end{figure}

Nevertheless, current VLN systems still lack a practical mechanism to model semantic progress. 
Modern Vision-Language-Action (VLA) navigation frameworks \cite{liunavid,cheng2024navila,wei2025streamvln} typically rely on end-to-end policy learning, where any notion of progress is implicitly buried within action prediction without an explicit representation. 
Traditional approaches \cite{zhu2020vision,zhu2020vision}, in contrast, approximate progress using numeric measures such as geometric ratios or remaining distance. 
However, these spatial proxies quantify movement but provide no reliable signal about the agent’s current position within the instruction, limiting their usefulness for decision making.
As a result, agents lack reliable semantic cues to connect past observations with task advancement, ultimately hindering coherent long-horizon behavior and interpretability.

\begin{figure*}[ht]
    \centering
    \includegraphics[width=1\linewidth]{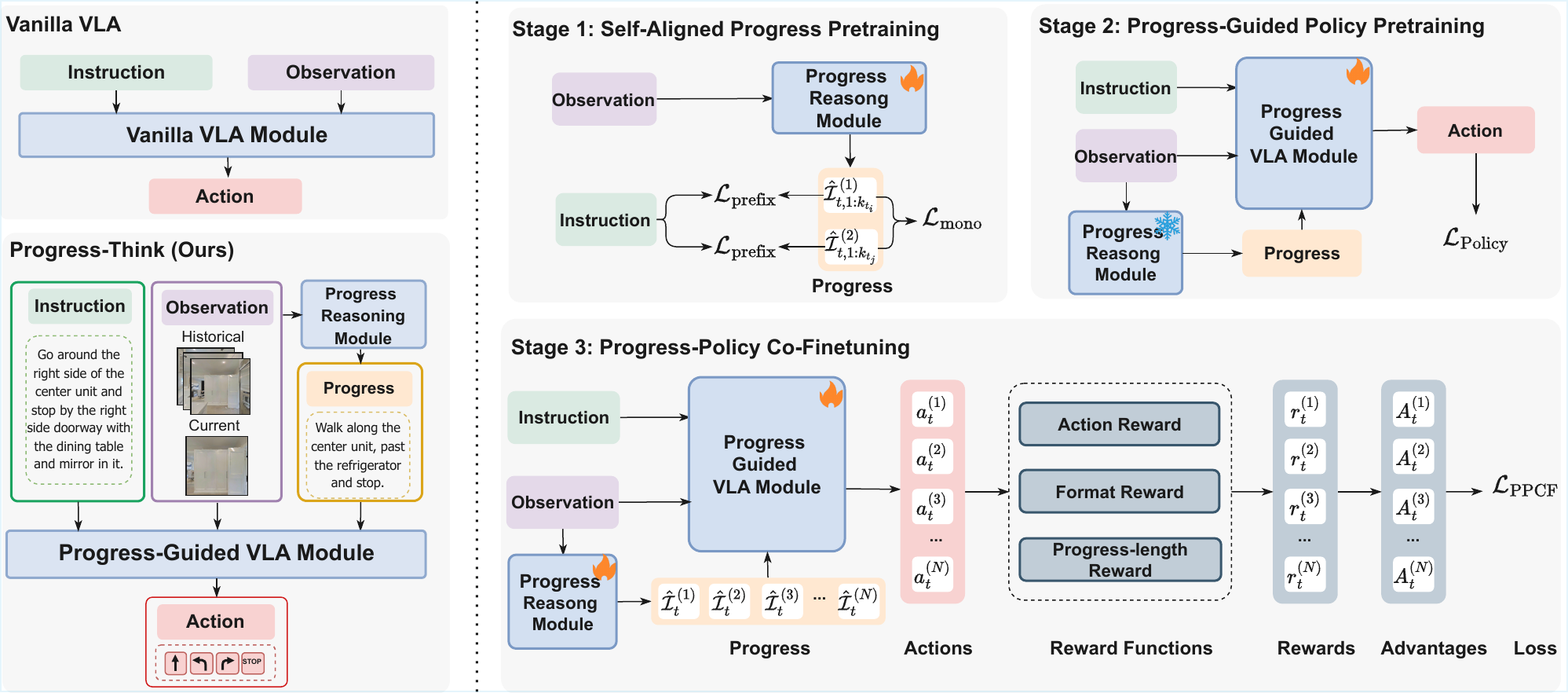}
    \caption{Overview of the Progress-Think framework and annotation-free training pipeline.
Compared with the vanilla Vision-Language-Action (VLA) model, Progress-Think introduces a Progress Reasoning Module to infer task progress and guide action generation. The model is trained in three stages: (1) Self-Aligned Progress Pretraining for progress pretraining with $\mathcal{L}_{\mathrm{SAPP}}=\mathcal{L}_{\mathrm{prefix}}+\mathcal{L}_{\mathrm{mono}}$, (2) Progress-Guided Policy Pretraining with frozen progress reasoning and supervised policy loss $\mathcal{L}_{\mathrm{Policy}}$, and (3) Progress-Policy Co-Finetuning, which jointly optimizes reasoning and policy through GRPO over groups of $N$ rollouts, using the objective $\mathcal{L}_{\mathrm{PPCF}}$. }
    \label{fig:main}
\end{figure*}

In this work, we present \textbf{Progress-Think}, a new perspective that models navigation progress through semantic progress reasoning.
Rather than estimating numeric completion, Progress-Think infers progress by leveraging the monotonic co-progression structure between accumulated observations and evolving instruction semantics, as shown in Fig.~\ref{fig:progress}, a critical property overlooked by prior methods.
The key idea is that progress emerges from how well the agent’s visual experience aligns with the instruction portion expected so far, revealing the agent’s true position in the task rather than its spatial displacement.
This semantic formulation enables the agent to track fulfilled and remaining instruction segments, supporting coherent behavior and yielding interpretable traces of instruction fulfillment.

Specifically, annotating semantic progress is prohibitively expensive, and no public datasets provide such supervision despite the input sequences’ inherent monotonic co-progression.
To achieve annotation-free semantic progress learning, Progress-Think proposes a three-stage formulation.
Self-Aligned Progress Pretraining first bootstraps a progress reasoning module through differentiable alignment between visual observations and instruction prefixes, enabling semantic progress inference without explicit labels.
Progress-Guided Policy Pretraining then uses the learned progress representation to supervise policy learning, aligning actions with the remaining instruction semantics for consistent long-horizon behavior.
Finally, Progress-Policy Co-Finetuning jointly refines the progress reasoning and navigation policy modules via reinforcement learning with a progress-aware reward, improving progress estimation and decision reliability.

Our main contributions are threefold:
\begin{itemize}
\item We first introduce semantic progress reasoning for VLN, reformulating progress estimation as stepwise visual–semantic alignment within VLA models.
\item We present an annotation-free three-stage framework that couples progress reasoning with policy learning through self-aligned progress pretraining, progress-guided policy adaptation, and progress–policy co-refinement.
\item We achieve state-of-the-art performance on R2R-CE and RxR-CE, with consistent improvements in success rate, path efficiency, and reasoning interpretability.
\end{itemize}

\section{Related Work}

Vision-language navigation~\cite{nguyen2019vision,wang2022towards,wu2020towards,schumann2024velma} lies at the intersection of vision, language, and action learning. To position our work within this context, we review two highly relevant research directions as follows.

\subsection{VLN with Large Pretrained Models}

Large-scale pretrained vision-language models (VLMs)~\cite{bai2025qwen2,liu2024nvila} have recently transformed the VLN landscape by providing stronger visual grounding and language understanding capabilities, and have further evolved into VLA models that couple perception, language, and control for embodied reasoning.
Compared to earlier navigation systems that relied on geometric mapping \cite{wang2023communication,wang2023distributed,li2023colslam}, recent VLA-based VLN systems~\cite{liunavid,zhang2024uni,cheng2024navila,wei2025streamvln} adopt monocular RGB inputs and perform supervised fine-tuning on pretrained multimodal backbones, yielding substantial gains in generalization and transferability.

To further enhance reasoning capabilities, several works incorporate Chain-of-Thought (CoT) reasoning supervision, including auxiliary reasoning tasks~\cite{wang2025auxthink} and slow-thinking pipelines~\cite{liu2025nav}.
MapDream \cite{lian2026mapdream} generates navigation-related maps as content for visual and geometric thinking, serving as the context for navigation VLA.
Progress reasoning, in contrast to CoT reasoning, better aligns the reasoning process with measurable task dynamics and advancement rather than linguistic plausibility, thereby establishing a tighter connection between reasoning and navigation.
Moreover, existing approaches depend heavily on external reasoning annotations, either generated by large VLMs~\cite{bai2025qwen2,achiam2023gpt,comanici2025gemini} or produced by human annotators, which are costly to obtain and may suffer from inconsistencies or biases, resulting in unreliable supervision in visually ambiguous or diverse navigation environments.

In contrast, our approach learns such progress-aware reasoning in a self-supervised manner directly from navigation interactions and integrates it with policy optimization, enabling coherent long-horizon decision-making without external reasoning labels.

\subsection{Progress Estimation in Embodied Navigation}

Progress estimation has been widely studied to mitigate error accumulation in long-horizon navigation.  
Recent VLA-based navigation systems typically encode historical visual observations as context \cite{zhang2024navid,cheng2024navila,zhang2024uni,wang2025think,wang2025monodream}, allowing the model to implicitly ground the current state to previously seen scenes.
Although this design helps the agent localize itself within the trajectory \cite{wang2025mambavo,wang2024gslamot}, the progress signal remains entangled with visual context rather than explicitly aligned with instruction semantics.
Other approaches add auxiliary objectives that regress a global completion ratio or related signals to assist policy learning \cite{zhu2020vision,ma2019regretful}. 
Milestone-based methods \cite{song2022one} aim for finer-grained control by identifying intermediate subgoals, but they typically require additional structure or supervision to produce step-level signals.

Despite these advances, prior progress representations are often either implicit, coarse, or learned independently from the action policy, which limits their ability to provide step-level semantic alignment between observations and instructions.  
In contrast, we propose a progress reasoning mechanism that infers detailed semantics from instruction–observation interactions and is explicitly coupled with policy optimization, improving accuracy and robustness.

%% file: sec/2_method.tex
\section{Method}
\label{sec:method}

\textbf{Overview.} 
We study monocular Vision-and-Language Navigation in Continuous Environments (VLN-CE), where an agent navigates in realistic spaces based on natural language instructions.
At each step $t$, the agent receives three types of inputs: the natural language instruction $\mathcal{I}$, the current egocentric RGB observation $o_t$, and a sampled history of past observations $\mathcal{O}_t$. 
The primary objective is to predict the next navigation action $a_t$ based on these multimodal inputs.

We propose Progress-Think, a progress-aware reasoning framework that decouples VLN into two complementary components, as illustrated in \cref{fig:main}. 
The Progress Reasoning Module reasons the agent’s executed progress. To learn this capability without sub-goal annotations, we introduce Self-Aligned Progress Pretraining, a self-supervised alignment objective that captures fine-grained progress semantics from partially matched trajectory segments. The predicted progress acts as a semantic anchor, dynamically guiding the Progress-Guided VLA Module to focus on the currently relevant subgoal. Finally, we design Progress-Policy Co-Finetuning, a reinforcement learning scheme that jointly improves progress reasoning and navigation policy learning.

\subsection{Self-Aligned Progress Pretraining}

A key challenge in progress reasoning is the lack of explicit supervision: existing VLN datasets provide only high-level instructions and final goals, without fine-grained step-level or sub-instruction annotations.
This makes it difficult for the agent to infer how much of the instruction has been completed from visual observations.
Moreover, acquiring large-scale progress annotations is prohibitively expensive, and labels generated by human or large foundation models tend to introduce noise, bias, and inconsistency.
To address this, we introduce Self-Aligned Progress Pretraining (SAPP), a self-supervised stage that bootstraps progress reasoning by deriving explicit supervision directly from the instruction itself through its inherent sequential structure.

We design a Progress Reasoning Module (PRM) $F_{P}$ to estimate the agent’s finished progress given the observed sequence of visual experiences.
At each step, with the current and historical visual observations as input, PRM predicts the portion of the instruction the agent has completed so far.
Formally, given the observation history $\mathcal{O}_{t} = \{o_{0}, \dots, o_{t-1}\}$, the module predicts the progress for step $t$:
\begin{equation}
\hat{\mathcal{I}}_{t} = \mathbf{F}_{P}(\mathcal{O}_{t}, o_{t}),
\end{equation}
where $\hat{\mathcal{I}}_{t}$ denotes the predicted semantic language span representing completed progress, which is fed as input to the VLA policy for subsequent action prediction.

Instead of relying on unavailable step-level annotations, SAPP derives explicit self-supervision directly from the instruction by leveraging two natural properties:
(1) a well-followed observation sequence should correspond to a prefix of the instruction executed so far, and
(2) progress should grow monotonically along the evolving visual observations.
Based on this, we propose Prefix-Subset Soft Cross-Entropy Loss and Monotonic Ordering Loss to achieve self-supervised progress learning.

We treat instruction prefixes as latent progress states and estimate a soft distribution over how much of the instruction has been completed.
We convert the decoder logits into a soft alignment distribution over prefix lengths $k \in [1, |\mathcal{I}|]$ for step $t$:
\begin{equation}
p_{\theta}(k \mid \mathcal{O}_{t}, \mathcal{I}) \propto
\exp\!\left(-\frac{\mathrm{CE}(\hat{\mathcal{I}}_{t,\,1:k}, \mathcal{I}_{1:k})}{\tau}\right),
\end{equation}
where $\tau$ is a temperature parameter, and CE is the cross-entropy function. 
A differentiable progress prefix length is defined as the distribution expectation:
$\hat{k}_{t} = \mathbb{E}_{p_{\theta}}[k]$.

This converts execution progress into a continuous and learnable semantic representation.
The Prefix-Subset Soft Cross-Entropy Loss is defined as:
\begin{equation}
\mathcal{L}_{\mathrm{prefix}} =
\mathbb{E}_{t}\Bigg[
-\tau \log \sum_{k}
\exp\!\left(
-\frac{\mathrm{CE}(\hat{\mathcal{I}}_{t,\,1:k}, \mathcal{I}_{1:k})}{\tau}
\right)
\Bigg].
\end{equation}

Since progress should evolve monotonically with the visual observation sequence, an earlier timestep should correspond to a prefix of a later one.
Formally, for two states sampled from the same episode where $t_{i} < t_{j}$, the predicted progress positions should satisfy $\hat{k}_{t_{j}} \ge \hat{k}_{t_{i}}$.
To enforce this structural property, we introduce a Monotonic Ordering Loss:
\begin{equation}
\mathcal{L}_{\mathrm{mono}} =
\mathbb{E}_{(i,j):\, t_{i} < t_{j}}
\left[
\max(0,\,  \hat{k}_{t_{i}} - \hat{k}_{t_{j}})
\right],
\end{equation}

The full training objective of Self-Aligned Progress Pretraining for the Progress Reasoning Module is:
\begin{equation}
\mathcal{L}_{\mathrm{SAPP}} =
\mathcal{L}_{\mathrm{prefix}} + \mathcal{L}_{\mathrm{mono}}.
\end{equation}

\subsection{Progress-Guided Policy Pretraining}

We design a Progress-Guided VLA (PG-VLA) policy that conditions action prediction on the estimated progress, as the Progress Reasoning Module infers semantic task completion progress that serves as an explicit conditioning cue for policy optimization, enabling coherent long-horizon decision making.
Given the predicted navigation progress from the PRM, PG-VLA $\pi_\theta$ generates navigation actions.
 At step $t$, the inputs include the instruction $\mathcal{I}$, the current observation $o_t$, the sampled history $\mathcal{O}_t$, and the navigation progress $\hat{\mathcal{I}}_t$ from PRM. Progress-Guided VLA Module $\pi_\theta$ predicts the navigation steps:
\begin{equation}
a_{t:t+K-1} = \mathbf \pi_\theta(\mathcal{O}_t, o_t, \mathcal{I}, \hat{\mathcal{I}}_t)
\end{equation}
where $a_{t:t+K-1}$ denotes the predicted $K$ action steps.

During Progress-Guided Policy Pretraining, the PRM is frozen to ensure stable guidance. We train the PG-VLA with a standard cross-entropy objective:
\begin{equation}
\mathcal{L}_{\mathrm{policy}}
=
- \log \mathbf \pi_\theta(  a_{t:t+K-1}^* | \mathcal{O}_t, o_t, \mathcal{I}, \hat{\mathcal{I}}_t)
\end{equation}
where $a_{t:t+K-1}^*$ denotes the ground-truth next $K$ steps.

\subsection{Progress-Policy Co-Finetuning}

Building upon the above two stages, we further enhance decision quality by jointly refining progress reasoning and policy. 
The progress estimates, initially learned from self-supervised alignment signals, benefit from being calibrated against task-level navigation objectives, leading to more coherent behaviors.
To jointly optimize the \textit{PRM} and \textit{PG-VLA},
we introduce Progress-Policy Co-Finetuning (PPCF), where progress reasoning and policy learning are optimized together through the reward signals derived from navigation outcomes. We adopt the Group Relative Policy Optimization (GRPO) framework \cite{shao2024deepseekmath}.

Given the predicted progress $\hat{\mathcal{I}}_t$ and the action sequence $a_{t:t+K-1}$, we define three complementary reward terms and sum them as the final reward in PPCF. Specifically, the \emph{action reward} enforces stepwise correctness, the \emph{format reward} guarantees syntactically valid action outputs, and the \emph{progress-length reward} regularizes the predicted progress to avoid premature or over-extended completion.

\noindent\textbf{Action Reward.}
We reward only the \emph{longest correct prefix} of the predicted action sequence:
\begin{equation}
r_{\mathrm{act}}
= \sum_{i=0}^{K-1} \prod_{j=0}^{i} \mathds{1}\big[a_{t+j} = a^*_{t+j}\big],
\end{equation}
where $\mathds{1}[\cdot]$ is the indicator function.  
The reward takes values in $\{0,1,2,...,K\}$: once the first incorrect action appears, all following steps receive no credit.

\noindent\textbf{Format Reward.}
We validate whether the predicted action sequence satisfies the required action syntax defined in Sec.~\ref{sec:action}:
\begin{equation}
r_{\mathrm{fmt}} =
\begin{cases}
1, & \text{if } a_{t:t+K-1} \text{ is in valid format},\\
0, & \text{otherwise}.
\end{cases}
\end{equation}

\noindent\textbf{Progress-Length Reward.}
Since the predicted progress text is a \emph{partial} sub-instruction, its length should naturally be shorter than or equal to the full instruction. To prevent premature or over-optimistic progress predictions, we apply a length constraint to penalize over-generation:
\begin{equation}
r_{\mathrm{len}} =
\begin{cases}
1, & \text{if } |\hat{\mathcal{I}}_t| \le |\mathcal{I}|,\\[3pt]
-\beta\big(|\hat{\mathcal{I}}_t|-|\mathcal{I}|\big), & \text{otherwise,}
\end{cases}
\end{equation}
where $\beta>0$ is a penalty factor.

\noindent\textbf{Final reward.}
The three terms are directly summed to form the final reward:
\begin{equation}
r_t = r_{\mathrm{act}} + r_{\mathrm{fmt}} + r_{\mathrm{len}}.
\end{equation}
This design simultaneously enforces stepwise action correctness, valid action structure, and well-calibrated progress estimation.

\noindent\textbf{Co-Refinement Optimization.}
During each rollout, the Progress Reasoning Module samples $N$ progress hypotheses
$\{\hat{\mathcal{I}}_t^{(1)}, \dots, \hat{\mathcal{I}}_t^{(N)}\}$,  
each of which conditions the Action Module to produce
$\{a_{t:t+K-1}^{(1)},..., a_{t:t+K-1}^{(N)}\}$. 
The advantage of each rollout $n$ is computed as:
\begin{equation}
A^{(n)} = \frac{r^{(n)} - \mathrm{mean}(r_t)}{\mathrm{std}(r_t)}.
\end{equation}

Both modules are updated together by maximizing the following GRPO-like objective:
\begin{equation}
\mathcal{L}_{\mathrm{PPCF}} =
- \mathbb{E}_n \left[
\min\left(
\rho^{(n)} A^{(n)},
\mathrm{clip}(\rho^{(n)}, 1 - \epsilon, 1 + \epsilon) A^{(n)}
\right)
\right],
\end{equation}
where
\begin{equation}
\rho^{(n)} =
\frac{\mathbf \pi_\theta(a_t^{(n)} \mid \mathcal{O}_t, o_t, \hat{\mathcal{I}}_t^{(n)})}
     {\mathbf \pi_{\theta_\mathrm{old}}(a_t^{(n)} \mid \mathcal{O}_t, o_t, \hat{\mathcal{I}}_t^{(n)})}
\cdot
\frac{\mathbf F_p(\hat{\mathcal{I}}_t^{(n)} \mid \mathcal{O}_t, o_t)}
     {\mathbf F_{p_\mathrm{old}}(\hat{\mathcal{I}}_t^{(n)} \mid \mathcal{O}_t, o_t)}.
\end{equation}

PPCF allows progress estimation and policy learning to mutually reinforce each other:
the PRM learns to produce progress representations that better match navigation objectives,
while the PG-VLA becomes more sensitive to progress-aware guidance, enabling coherent long-horizon navigation.

\section{Implementation Details}

\subsection{Dataset}
We construct training data using continuous VLN simulators from the training splits of R2R-CE \cite{krantz2020beyond}, RxR-CE \cite{ku2020room}, and ScaleVLN \cite{wang2023scaling,wei2025streamvln}. 
Their annotated trajectories are converted into step-level samples, yielding 1200K state–action pairs respectively. 
To support progress reasoning, we additionally generate partial-trajectory samples by pairing trajectory prefixes with full instructions, providing weak progress supervision without subgoal labels. 
Following the DAgger strategy \cite{ross2011reduction}, we further collect 500K non-oracle samples by executing exploratory rollouts in training scenes, improving robustness to off-distribution states. 

\subsection{Model Training}

We train Progress-Think in three stages, as illustrated in \cref{fig:main}. 
All training experiments are conducted on 8 NVIDIA H20 GPUs. 
Stage 1 takes approximately 8 hours, while stages 2 and 3 each take about 60 hours.

In the Self-Aligned Progress Pretraining, we initialize the Progress Reasoning Module from the pretrained NVILA-2B~\cite{liu2024nvila} weights and perform self-supervised progress learning. For each trajectory, the full instruction is used as the supervision target, enabling the model to acquire coarse semantic progress understanding from instruction–trajectory alignment before interacting with action supervision. This stage is trained for one epoch with a learning rate of $1\times10^{-5}$.

In the Progress-Guided Policy Pretraining, the PRM is frozen to provide stable progress guidance, and we train the PG-VLA, also initialized from NVILA-2B \cite{liu2024nvila}. The policy is trained using expert oracle trajectories together with DAgger-generated trajectories, optimized using cross-entropy loss over the next three action steps. This stage also runs for one epoch with a learning rate of $1\times10^{-5}$.

The third stage performs Progress–Policy Co-Finetuning, where PRM and PG-VLA are jointly optimized via reinforcement learning. We adopt a GRPO-style optimization scheme with a rollout size of 4, a KL coefficient of 0.0, and a clipping parameter of 0.28. This stage is trained for 3,000 steps with a learning rate of $1\times10^{-6}$, enabling progress reasoning and policy learning to co-adapt and improve long-horizon consistency.

\begin{table*}[htbp]
  \centering
  \caption{Comparison of different methods on the R2R-CE Val-Unseen split. Observations used include single RGB camera (S.RGB), depth sensor (Depth) and panoramic view (Pano.). $\dagger$ indicates methods without using LLMs. External data refers to sources beyond the navigation simulator, such as real-world web data, general VQA datasets, and other similar resources.}
  \label{tab:r2r_val_unseen}
  \resizebox{0.8\textwidth}{!}{
  \begin{tabular}{l ccc c c c c r}
    \toprule
    \multirow{2}{*}{Method}  & \multicolumn{3}{c}{Observation} & \multicolumn{4}{c}{R2R-CE Val-Unseen} & {Training} \\ 
    \cmidrule(lr){2-4} \cmidrule(lr){5-8} \cmidrule(lr){9-9}
     & {S.RGB} & {Depth} & {Pano.} & {NE $\downarrow$} & {OSR $\uparrow$} & {SR $\uparrow$} & {SPL $\uparrow$} & {External Data} \\
    \midrule
    BEVBert$^{\dagger}$\cite{an2022bevbert}        & & $\checkmark$ & $\checkmark$  & 4.57 & 67.0 & 59.0 & 50.0 &   - \\
    ETPNav$^{\dagger}$\cite{an2024etpnav}         & & $\checkmark$  & $\checkmark$  & 4.71 & 65.0 & 57.0 & 49.0 &   - \\
    ENP-ETPNav$^{\dagger}$\cite{liu2024vision}  & & $\checkmark$ & $\checkmark$  & 4.69 & {65}   & {58}   & {50}   &  - \\ 
    \midrule
    Seq2Seq$^\dagger$\cite{krantz2020beyond}      &  $\checkmark$ &$\checkmark$ & & 7.77 & 37.0 & 25.0 & 22.0 &  -  \\
    CMA$^\dagger$\cite{krantz2020beyond}      & $\checkmark$ &$\checkmark$ & & 7.37 & 40.0 & 32.0 & 30.0 &  -  \\
    LAW$^\dagger$\cite{ray2021language}    & $\checkmark$  &$\checkmark$ & & 6.83 & 44.0 & 35.0 & 31.0 &  -  \\
    CM2$^\dagger$\cite{georgakis2022cross}   & $\checkmark$ &$\checkmark$ & & 7.02 & 41.0 & 34.0 & 27.0 &  -  \\
    WS-MGMap$^\dagger$\cite{chen2022weakly}  & $\checkmark$ &$\checkmark$ & & 6.28 & 47.0 & 38.0 & 34.0 &   - \\
    sim2real$^\dagger$\cite{wang2024sim}   & $\checkmark$ & $\checkmark$ & & 5.95 & 55.8 & 44.9 & 30.4 &  -  \\ 
    NavMorph$^\dagger$\cite{yao2025navmorph} & $\checkmark$ & $\checkmark$ & & 5.75 & 56.9 & 47.9 & 33.2 &  -  \\ 
    NaVid-4D \cite{liunavid}     & $\checkmark$ & $\checkmark$ & & 5.99 & 55.7 & 43.8 & 37.1 & 1500K  \\
    \midrule
    Uni-NaVid\cite{zhang2024uni}    & $\checkmark$ & & & 5.58 & 53.3 & 47.0 & 42.7 &  2300K  \\
    NaVILA\cite{cheng2024navila}   & $\checkmark$ & & & \underline{5.22} & {62.5} & 54.0 & 49.0 & 2215K   \\
    Aux-Think \cite{wang2025think}  & $\checkmark$ & & & 5.49 & \underline{62.9} & 55.7 & 48.7 & 1600K  \\ 
    NaVid \cite{zhang2024navid}     & $\checkmark$ & & & 5.47 & 49.1 & 37.4 & 35.9 &   {0K}  \\
    ActiveVLN \cite{wang2025dynam3d}  & $\checkmark$ &   & & 5.34 & 62.1 & 52.9 & 45.7 & 0K \\
    MonoDream\cite{wang2025monodream}   & $\checkmark$ & & & 5.45 & 61.5 & \underline{55.8} & \underline{49.1} & {0K} \\
    Progress-Think (ours)   & $\checkmark$ & & & \textbf{4.68} & \textbf{63.6} & \textbf{60.1} & \textbf{53.6}  & 0K \\
    \bottomrule
  \end{tabular}}
\end{table*}

\subsection{Action Design}
\label{sec:action}

We follow the continuous VLN-CE setting and formulate navigation as a sequence of discrete actions.  
At each step, the predefined action space consists of \texttt{move forward}, \texttt{turn left}, \texttt{turn right}\}, and \texttt{stop}.
The forward action includes step sizes of 25 cm, 50 cm, and 75 cm, while the turn actions are parameterized by rotation angles of 15$^\circ$, 30$^\circ$, and 45$^\circ$.
Our Progress-Guided Action Module predicts next three action steps ($K=3$), following the multi-step prediction design in prior VLN works~\cite{zhang2024uni,wang2025auxthink}.

%% file: sec/3_experiment.tex
\section{Experiments}
\label{sec:experiments}

\subsection{Experiment Setup}
\label{sec:exp_set}

\subsubsection{Experimental environments} 
We evaluate our method on the VLN-CE benchmarks R2R-CE \cite{krantz2020beyond} and RxR-CE \cite{ku2020room} following the standard VLN-CE settings. 
All the methods are evaluated on the R2R-CE val-unseen split and RxR-CE val-unseen split.

\subsubsection{Metrics} We follow the standard VLN evaluation protocol \cite{krantz2020beyond, ku2020room} to evaluate navigation performance for all methods, including success rate (SR), oracle success rate (OSR), success weighted by path length (SPL), and the navigation error from the goal (NE). Among them, SR and SPL are widely regarded as the primary metrics, reflecting the task completion and path efficiency respectively.

\subsection{Comparison on VLN-CE Benchmarks}

We evaluate our method on standard VLN-CE benchmarks, including R2R-CE and RxR-CE, which provide continuous navigation environments in photorealistic 3D indoor scenes.
We first report results on the R2R-CE val-unseen split in \cref{tab:r2r_val_unseen}. To ensure fair comparison, methods are grouped based on sensor modality, and approaches that do not rely on large language models are marked with ($\dagger$).

As shown in \cref{tab:r2r_val_unseen}, Progress-Think achieves strong performance without using any external data beyond standard simulator supervision. This confirms the data efficiency of our approach, which stems from explicitly modeling navigation progress rather than relying on external reasoning labels. 
Remarkably, Progress-Think outperforms depth- and panorama-based methods while using only monocular RGB.
We attribute the performance gains to two key factors:
(1) Self-Aligned Progress Pretraining, which provides fine-grained alignment between partial trajectories and language instructions, enabling the agent to maintain an accurate sense of task completion;
(2) Progress-policy Co-Refinement Learning, where decision making is directly conditioned on structured progress representations, reducing ambiguity and mitigating error accumulation during navigation. Notably, these gains are achieved without increasing model scale or introducing external supervision, highlighting the effectiveness of progress reasoning.

\textbf{Unseen-Dataset Generalization.}
We assess the generalization capability of our method on the RxR-CE Val-Unseen split. Notably, all the methods are trained without any RxR-CE training data and our method achieves the state-of-the-art performance. This demonstrates the strong transferability of our framework.

\begin{table}[htbp]
\centering
\caption{Unseen-Dataset generalization performance on the RxR-CE Val-Unseen split. All results are obtained only training on the R2R-CE training set. }
\label{tab:rxr_results}
\scalebox{0.88}{
\begin{tabular}{@{}l c c c c@{}} 
\toprule
\multirow{2}{*}{Method} & \multicolumn{4}{c}{RxR-CE Val-Unseen} \\
 \cmidrule(lr){2-5}
& {NE $\downarrow$} & {OSR $\uparrow$} & {SR $\uparrow$} & {SPL $\uparrow$} \\
\midrule
Seq2Seq\cite{krantz2020beyond}  & 11.8            & 5.02            & 3.51            & 3.43            \\
CMA\cite{krantz2020beyond}      & 11.7            & 10.7            & 4.41            & 2.47            \\
LAW\cite{ray2021language} & 10.87           & 21.0            & 8.0             & 8.0             \\
CM2\cite{georgakis2022cross}   & 8.98            & 25.3            & 14.4            & 9.2             \\
WS-MGMap\cite{chen2022weakly} & 9.83            & 29.8            & 15.0            & 12.1            \\
A$^2$NAV\cite{chen20232}     & {-}             & {-}             & 16.8            & 6.3             \\ 
NaVid\cite{zhang2024navid}   & \underline{8.41}   & {34.5}  & {23.8}  & {21.2}  \\
MonoDream\cite{wang2025monodream}   & 8.57  &  \underline{35.9}  & \underline{25.1}   &  \underline{21.6} \\
Progress-Think (ours)   & \textbf{8.30}  &  \textbf{38.3}  & \textbf{27.5}   &  \textbf{22.7} \\
\bottomrule
\end{tabular}}
\end{table}

\begin{figure*}[ht]
    \centering
    \includegraphics[width=1\linewidth]{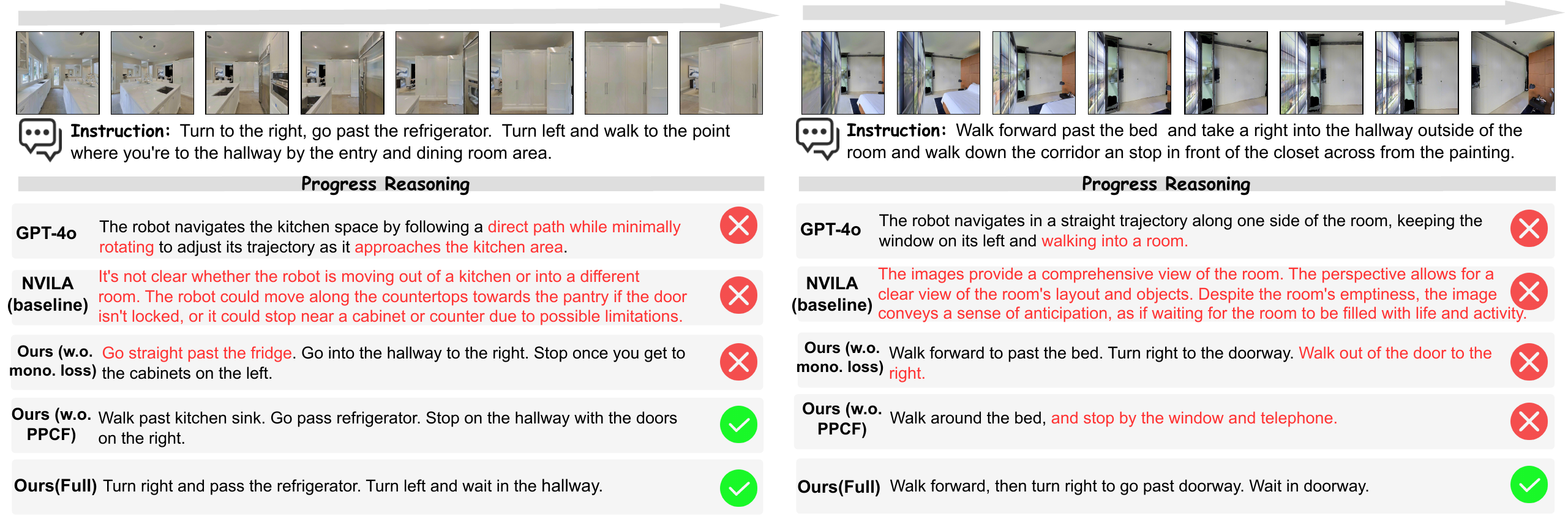}
    \caption{ Qualitative comparison of progress reasoning quality.
Across two representative scenes, we compare how different models infer navigation progress from historical observations. GPT-4o and NVILA\cite{liu2024nvila} often produce generic or instruction-misaligned descriptions and occasionally exhibit hallucinations, limiting their usefulness for tracking progress and making it difficult for the agent to align its behavior with the intended navigation steps. Our ablated variants (without monotonic loss or without Progress-Policy Co-Finetuning) capture partial progress but tend to be less consistent and concrete, leading to incomplete guidance. In contrast, the full Progress-Think model produces concise, instruction-style reasoning that adheres closely to the task and accurately reflects the agent’s evolving progress, enabling more coherent and reliable navigation.}
    \label{fig:demo}
\end{figure*}

\textbf{Qualitative comparison of progress reasoning.}
As shown in \cref{fig:demo}, GPT-4o and NVILA \cite{liu2024nvila} often produce instruction-agnostic descriptions that are weakly aligned in the visual trajectory, offering limited cues for understanding navigation progress. Removing monotonic orderding constraints or RL coupling yields partially correct but less structured and less stage-aware reasoning. In contrast, Progress-Think generates concise, step-aligned, and stage-aware reasoning that closely follows the instruction and reflects the agent’s actual progress.

\subsection{Comparison on Progress Variants}

To understand the impact of different progress representations, we compare two variants on the R2R-CE benchmark in Table~\ref{tab:semantic_vs_numeric}.
\emph{Numeric Progress Regression} predicts the completion percentage of the trajectory, following traditional scalar progress estimation used in prior VLN work \cite{zhu2020vision}.
\emph{Instruction Reasoning} trains the model to reconstruct the full instruction from the complete trajectory, following \cite{zhang2024navid,wang2025auxthink}. 

Semantic progress achieves the strongest performance among all variants. Numeric progress regression remains a scalar signal and lacks semantics, and instruction reasoning captures only a global summary of the trajectory. Our semantic progress formulation, by contrast, provides stepwise, instruction-style alignment that directly supports actionable decision-making in long-horizon navigation.

\begin{table}[htbp]
\centering
\caption{Comparison of different progress variants.}
\scalebox{0.9}{
\begin{tabular}{l|cccc}
\toprule
Method & NE$\downarrow$ & OSR$\uparrow$ & SR$\uparrow$ & SPL$\uparrow$ \\
\midrule
Numeric Progress Regression  & 8.25 & 37.7 & 33.4 & 26.2 \\
Instruction Reasoning  &\underline{7.67} & \underline{45.8}  & \underline{37.7} & \underline{32.1} \\
Progress-Think (ours) &\textbf{6.84} & \textbf{50.4} & \textbf{43.8} & \textbf{38.5} \\
\bottomrule
\end{tabular}}
\label{tab:semantic_vs_numeric}
\end{table}

\subsection{Ablation Study}
To enable fast and efficient validation of our proposed components, we conduct all ablation experiments on the R2R-CE benchmark.
\cref{tab:ablation_full_combined} presents the results of our ablation study on the proposed Self-Aligned Progress Pretraining (SAPP) and Progress-Policy Co-Finetuning (PPCF). We observe that introducing SAPP alone already leads to consistent improvements across all metrics compared to the baseline, indicating that explicitly reasoning about progress helps the agent better interpret the instruction and track navigation progress. Adding PPCF on top of SAPP further boosts the SR and SPL metrics, demonstrating that coordinating policy optimization with progress reasoning enables more accurate navigation. These results confirm that both components contribute positively.

\subsubsection{Impact of Self-Aligned Progress Pretraining}

\cref{tab:ablation_full_combined} shows that incorporating the Prefix-Subset Soft Cross-Entropy Loss improves NE, SR, and SPL, suggesting that soft supervision over instruction prefixes provides finer-grained guidance for estimating partial task completion. Adding the Monotonic Ordering Loss further boosts all metrics, indicating that enforcing temporal consistency across predicted progress prevents regressions and stabilizes learning. Together, these components enable the PRM to generate more accurate and reliable progress estimates, which translates into more efficient navigation.

\begin{table}[htbp]
\centering
\caption{Unified ablation study on SAPP and PPCF.
SAPP: Self-Aligned Progress Pretraining (Prefix: prefix-soft CE loss, Mono: monotonic ordering loss).
PPCF uses two reward configurations: AR+FR (Action Reward + Format Reward) and PLR (Progress-Length Reward).}
\scalebox{0.87}{
\begin{tabular}{cc|cc|cccc}
\toprule
\multicolumn{2}{c|}{SAPP Losses} &
\multicolumn{2}{c|}{PPCF Rewards} &
\multicolumn{4}{c}{Metrics} \\
Prefix & Mono & AR+FR & PLR &
NE$\downarrow$ & OSR$\uparrow$ & SR$\uparrow$ & SPL$\uparrow$ \\
\midrule
  &  &  &  & 8.16 & 44.1 & 33.0 & 28.3 \\
\checkmark &  &  &  & 7.26 & 45.8 & 39.4 & 34.6 \\
\checkmark & \checkmark &  &  & 6.94 & 48.4 & 41.4 & 36.5 \\
\checkmark & \checkmark & \checkmark &  & 7.16 & 46.2 & 40.0 & 35.0 \\
\checkmark & \checkmark & \checkmark & \checkmark &
\textbf{6.84} & \textbf{50.4} & \textbf{43.8} & \textbf{38.5} \\
\bottomrule
\end{tabular}}
\label{tab:ablation_full_combined}
\end{table}

\subsubsection{Impact of Rewards in Co-Finetuning}
The results in \cref{tab:ablation_full_combined} also demonstrate the effectiveness of the Progress-Length Reward in Progress-Policy Co-Finetuning.
Introducing this reward brings consistent improvements across all metrics, most notably reducing NE and increasing SR and SPL.
This indicates that constraining progress predictions to realistic instruction lengths prevents over-estimation of task completion, yielding more reliable progress supervision and more accurate long-horizon decision-making.
The overall gains confirm that explicitly regularizing progress not only stabilizes learning but also enhances navigation efficiency and goal attainment.
\begin{table}[htbp]
\centering
\caption{Ablation on executed action steps. }
\scalebox{1}{
\begin{tabular}{c|cccc}
\toprule
\#Steps & NE$\downarrow$ & OSR$\uparrow$ & SR$\uparrow$ & SPL$\uparrow$ \\
\midrule
1  & 4.73 & 62.1 & 58.0 & 51.1 \\
2  & 4.80 & 62.2 & 58.9 & 52.3 \\
3  & \textbf{4.68} & \textbf{63.6} & \textbf{60.1} & \textbf{53.6} \\
\bottomrule
\end{tabular}}
\label{tab:action}
\end{table}
\subsubsection{Impact of Different Executed Action Steps}

To assess how many actions should be executed per prediction, we compare different execution lengths while keeping the PG-VLA prediction horizon fixed to three actions. As shown in Table \ref{tab:action}, executing only one action leads to short-sighted behavior and frequent re-planning, resulting in higher navigation errors.
Executing all three predicted actions achieves the best performance across all metrics, suggesting that longer execution enables more stable decision-making and better use of progress-aware guidance.

\subsection{Robustness to Trajectory Granularity}
\begin{figure}[h]
    \centering
    \includegraphics[width=0.6\linewidth]{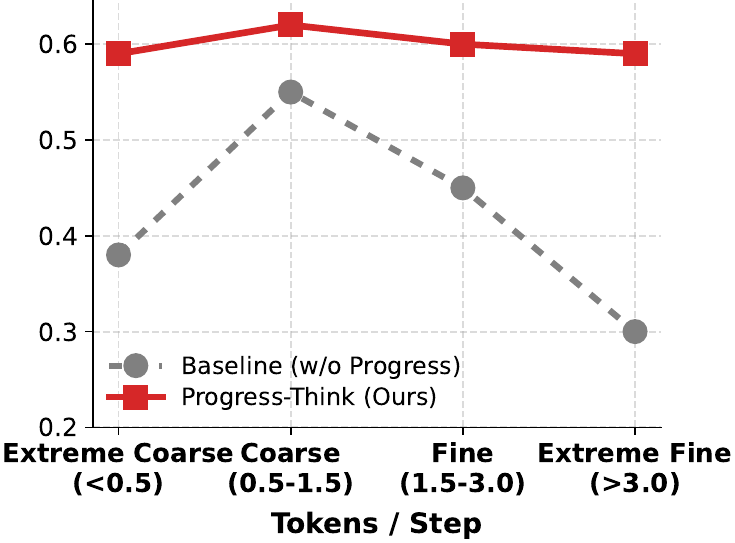}
    \caption{Granularity analysis on R2R-CE.}
    \label{fig:line}
\end{figure}

Our method is robust to weak instruction-trajectory length correlation, as demonstrated by the analysis on granularity (instruction length divided by trajectory length) in \cref{fig:line}. Crucially, in the granularity analysis, our method achieves the largest gains for extremely coarse or extremely fine instructions.
\vspace{-0.43cm}
\subsection{Model Size and Efficiency}

To assess the computational footprint of Progress-Think, we compare its parameter scale and inference efficiency with representative VLN approaches. All methods are evaluated under the same hardware setting using Nvidia 4090 GPU to ensure fairness. The efficiency is compared as the average time required to complete one successful episode on the R2R-CE. As shown in \cref{tab:model_efficiency}, Progress-Think achieves less inference speed while maintaining a moderate model size, demonstrating that its progress reasoning and policy refinement mechanisms introduce minimal overhead. This balance of accuracy and efficiency makes Progress-Think suitable for long-horizon navigation scenarios.

\begin{table}[htbp]
\centering
\caption{Comparison of model size and inference efficiency on R2R-CE val-unseen. }
\scalebox{0.9}{
\begin{tabular}{l|cc|c}
\toprule
Method & Params$\downarrow$ & Time / Episode$\downarrow$ & SR$\uparrow$ \\
\midrule
NaVILA \cite{cheng2024navila}     & 8B & 56.84s  & 54.0 \\
Aux-Think \cite{wang2025auxthink} & 8B & 58.04s &  54.8\\
\midrule
\textbf{Progress-Think (ours)} & \textbf{2B+2B} & \textbf{36.60s} & \textbf{60.1} \\
\bottomrule
\end{tabular}}
\label{tab:model_efficiency}
\end{table}

\section{Conclusion}

In this paper, we presented \textbf{Progress-Think}, a semantic progress reasoning framework for Vision-Language Navigation in continuous environments.
Progress-Think is driven by the insight that VLN inputs exhibit an inherent monotonic co-progression, where accumulating visual inputs advance the aligned instruction prefix, a structural property overlooked by existing methods.
Through a three-stage optimization pipeline, our method enables annotation-free progress reasoning and more consistent decisions.
Experiments on R2R-CE and RxR-CE demonstrate clear gains in navigation performance, temporal coherence, and interpretability.
We hope this work encourages tighter integration of progress reasoning and policy learning in embodied agents, making semantic progress a central element of robust long-horizon navigation.